\documentclass[a4paper,10pt]{article}
\usepackage{times}
\usepackage[dvips]{epsfig}

\begin{document}


\title{Building a linguistic corpus from bee dance data}

\author{Dr. J.J. Paijmans \\
Department of Linguistics and Artificial Intelligence\\
Tilburg University\\
Tilburg, Netherlands \\
j.j.paijmans@uvt.nl}

\maketitle

\thispagestyle{empty}

\abstract{This paper discusses the problems and possibility of collecting bee
dance data in a linguistic {\em corpus} and use linguistic instruments such as
Zipf's law and entropy statistics to decide on the question whether the dance
carries information of any kind. We describe this against the historical
background of attempts to analyse nonhuman communication systems.\\
}

\section{Introduction}
The idea for this paper originated from a small paper that my daughter wrote
for her bachelor study biology at the University of Wageningen,
Netherlands. In this paper she discussed the views on the function of the
so-called 'honey bee dance', more in particular the opposition of A. Wenner
\cite{wenner:2002,wenner/wells:1973} and others against the established theory
first put forward by K. von Frisch\cite{Frisch:1947} (non vidi) that
the shape and direction of a dance performed by a bee communicate to other
bees the direction and distance of a food source (for a
recent discussion of the established views see \cite{Dyer:2002}).

As a computerlinguist I am not qualified to judge between the two
views. Of course the 'bee dance' is often referred to in linguistic
textbooks as an example of non-human communication, although not many
linguists would go so far as to call it 'speech' or even 'language'.

However, I wondered if the techniques that are used in te field
of {\em corpus linguistics} could be applied to the data that were
collected by the entymologists in studying the bee dance. More in
particular: if it could be demonstrated that the data in this corpus
had certain features in common with linguistic corpora, this could
indicate that the bees communicated {\em something}. Whether this
'something' concerned the location of a particular succulent brand
of honey or just local hive gossip was (to me) a matter of no concern.

It rapidly became clear that there already existed ample literature on the
subject of animal communication. Indeed the first attempt to statistically
analyse bee dance data in the light of Shannon's information theory is from
Haldane and Spurway in 1954 \cite{haldane:1954}, who used the original data
of von Frisch. Also, the theories of Zipf, notably Zipf's laws and the
principle of last effort played an important role in the analysis of both
human language and animal communication systems and even in the study of
manuscripts of unknown origin such as the Voynich manuscript\footnote{The
Voynich manuscript is a 16th century manuscript written in an unknown
language and alphabet, but probably a hoax}\cite{landini:2000,landini:2001}.


Often the debate on whether to call a certain communication system a
'language' is based on different definitions of 'language' or even
philosophical leanings of the participants in the debate. This is even
true in the Wenner attacks on the theory of von Frisch. Als Kak \cite{kak:1991} remarks: ``It
appears that the controversy is partly of a semantic nature. What does
language mean? (...)  Operationally this means that a language must be
associated with a vocabulary of basic signs and sounds and a grammar
that allows the signs or sounds {\em to be combined into an unlimited
number of statements}.''(my emphasis, P.). This key
notion of language having an {\em unlimited} number of elements
returns in many papers (e.g. Ujhelyi \cite{Ujhelyi:1996}) and is also
true for the words in human language, as was mathematically proven by
Kornai\cite{kornai:2002}. Ujhelyi also points out that the main difference
between human and animal communication is that animal
communication only allows for a limited set of messages, which are in
general genetically fixed. However, the work of McCowan and her
collaborators\cite{Cowan/Hanser/Doyle:2002,mccowan:1999} at least
proves that enough variation exists in the communicaton of humans,
dolphins and squirrel monkeys to observe Zipf's law at work.

I started this research in the hope that enough data could be obtained from
bee dance data to also observe Zipf's law in action, but it turned out
that this kind of data was not at all suitable for this kind of analysis.

Before data can be analysed it must be collected and stored in a suitable
format. After the explanation of the principles of the honey dance, and the
mathematical principles underlying the Zipfian laws and entropy, we will
proceed to the problems of collecting the data and the analysis or
translation to a format that is suitable for comparing the patterns with those
of other (human) languages.

\section{Animal communication}

\subsection{The bee dance}

\begin{figure}
\begin{center}
\epsfig{file=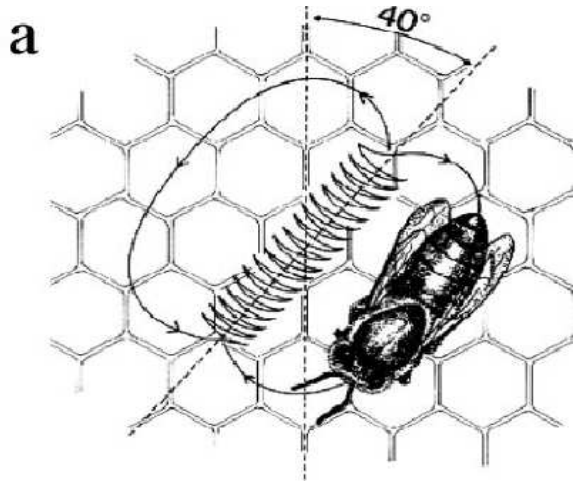,height=2.5in,width=3in}
\caption{\em Honey bee dance (from \cite{Dyer:2002})}
\label{beedance}
\end{center}
\end{figure}

In 1947 it was observed by K. von Frisch \cite{Frisch:1947} that there was a
certain pattern to the movements that a bee makes on the comb (see figure
\ref{beedance}). This pattern described approximatly a figure 8, and on the
traverse the bee waggles its abdomen as if for emphasis. Von Frisch observed
that the angle of this traverse with the vertical indicated the direction of a
honey source (more precise: the angle with the vertical correlated
with the angle vis-a-vis the sun of the honey source). The duration of the
dance, then, would indicate the distance.

Von Frisch and his followers also noted that other bees indeed seemed
to observe the dance closely, and afterwards acted upon it, therefore
they assumed that the dance had a communicative function and the bee
dance theory was born.  Because of the exciting nature of this
discovery, many authors followed up and meticulously analized every
possible aspect of the dance. We already mentioned the controversy
generated by Wenner and Wells and the seminal work of Halldane and
Spurway, but we could add dozens of scientists. Haldane and Spurway
computed the correct amount of information in a message for
non-discrete values such as direction, coming to approx. 5 cybernetic
units (sic!) as to direction, 4 to 5 as to distance and 2 to 3 as to
the number of workers needed. This totals to about 12 bits, equivalent
to a human language of 4000 phrases (signifiants with corresponding
signifi\'es), needing less than a hundred words by human or english
standards. Put differently, a code of all possible combinations of
only three characters would cover the communication system of the
honey bee dance. Towne and Gould \cite{Towne:1988} were among the many
scientists who continued research in the spatial precison of this
communication, giving much attention to the mathematics of
communication and survival in circumstances where the scatter and
quantity of the flowers varied. I was mildly amazed that I could not
find the observation radius of the recruit in flight as a factor in
the discussion.

A typical database for the processing of bee dance data might look
like table~\ref{tabel1}. Here, every observation includes three
estimates of the angle with the vertical direction of the central line
of the '8'. From this angle and the height of the sun at that moment,
the direction is computed. For that reason, the azimuth and
time-of-day are also included in the table.  The distance of the honey
source is deducted from the duration of the dance; for this purpose
the number of dances, the total time and the average time are noted in
the table. Finally the data are translated to an X and Y value for
subsequent plotting.

\begin{table}

\begin{footnotesize}
\begin{center}
\begin{tabular}{|l|l|l| |l|l|l| |l|l|l|l|l|}
\hline
     &  Ang.  &       &         &  Dir.   &         &  Av      &  no.     &   Dur.   &  mm:ss  \\
     &        &       &         &         &         &  Dir.    &  dances  &   Total  &  Average\\
\hline
195  &  195   &  200  &  346.4  &  346.4  &  351.4  &  348.07  &   7  &  00:20.40  &  00:02.91\\
200  &  200   &  200  &  354.1  &  354.1  &  354.1  &  354.10  &  10  &  00:36.02  &  00:03.60\\
  5  &   10   &   10  &  159.5  &  164.5  &  164.5  &  162.83  &  10  &  00:35.50  &  00:03.55\\
...  &  ...   &  ...  &  ...    &  ...    &  ...    &  ...     &  ... &  ...       &  ...\\
\hline
\end{tabular}
\begin{verbatim}

\end{verbatim}

\begin{tabular}{|l|l|l|l|l|l|l|l|l|l|}
\hline
Time     &  Az.    &  Dist.  &  Rad. &  cos   &  sin   &  X      &  Y      &  Pollen\\
hh:mm    &         &   [km]  &       &        &        &         &         &        \\
\hline
10:03    &  151.4  &  0.75   &  6.1  &  0.98  &  -0.21 &  -0.16  &   0.74  &  Pollen\\
10:10    &  154.1  &  1.32   &  6.2  &  0.99  &  -0.10 &  -0.14  &   1.31  &  Pollen\\
10:11    &  154.5  &  1.26   &  2.8  &  -0.96 &   0.30 &   0.37  &  -1.21  &  Pollen\\
...      &  ...    &  ...    &  ...  &  ...   &  ...   &  ...    &  ...    &  ...\\
\hline
\end{tabular}
\end{center}
\caption{A series of observations of the bee dance (courtesy M. Beekman).}
\label{tabel1}

\end{footnotesize}
\end{table}

\subsection{Other examples}

It is tempting to consider the human communication system as a
descendant of evolutionary earlier systems such as, indeed, the bee
language, but of course this is not necessarily true. On the contrary,
it seems that the communication of humans and higher mammals are based
on sound and ambiguity\cite{Cowan/Hanser/Doyle:2002}. The direct
predecessor of human language should be found in territoriality
messages and monogamous duetting \cite{Ujhelyi:1996}. As we will see
below, it conforms to Zipfian laws and is firmly rooted in
mathematics. Other animals communicate over a variety of channels,
including sound, but also in movement, smell such as our honey
bees. Haldane and Spurway put forward the notion that the honey bee
dance is a higly ritualized {\em intention movement}.

Phylogenetically quite near the honey bee we find other communication, e.g. of
ants. Reznikova\cite{Reznikova/Ryabko:2001} notes that the duration of the
contact between scout and recruits is linearly correlated to the number of the
travese where food was found, and suggests that this is a indication that ants
can count. In our opinion this resembles more a playback-like report than an
indication of counting discrete units.

Just the observation of an act of the scouting bee and the subsequent
reaction of the recruits is not enough to call the behaviour
'language' or 'communication', even if the reaction of the recruits
makes sense in the context. To give an example: imagine a student
entering his dormitory, smelling of beer and staggering around in
circles before collapsing in a chair. His friends would observe this
behaviour, come to the conclusion that at least one pub in the
vicinity had opened its doors and walk out to see if they can find a
place where music and light indicate the presence of an open pub.  I
would hesitate to call the action of the first student
'communication', and certainly not that particular communication that
is called language. If different pubs would sell different beers, that
caused different reactions in the 'scout student' so that his friends
would not only recognize the fact that a pub had opened, but also {\em
which} pub had opened, this still would not be called 'language',
because language presupposes intentionality\cite{kak:1991}. But
substituting one problem of the meaning of 'language' by that of the
meaning of 'intentionality' does not really help; what we need is a
model where 'language' is defined and inbedded within the broader term
of communication. As we will see below, thanks to the work of Ferrer
and Sol\'e, this is possible within the general framework of Zipf's
laws and the principle of least effort.


\section{Least effort and mathematics}

Zipf's law is the observation that frequency of occurrence of some
event $P$ as a function of the rank $i$, where the rank is determined
by the above frequency of occurrence, is a power-law function $Pi
\approx 1/i^{a}$ with the exponent $a$ close to unity. This is true
for interesting phenomena such as the frequencies of words in human
languages, and for the size of the population of cities or the
division of wealth. Later other researchers such as Wentian Li
\cite{li92random} have proven that Zipf's law also holds for less
interesting phenomena, such as randomly generated sequences of
characters. Research like that of Cohen, Mantegna and
Havlin\cite{havlin-can} tried to find the differences between such
'random languages' (my terminology\footnote{The original term in the
paper is 'artificial', but this causes problems with the terms used
for computer languages and such like.}) and real languages.  In the
paper mentioned here, it was found that the value of the
Zipf-derivates and especially the entropy of the word frequencies
differed after all between the natural language texts and the
artificial texts. Landini \cite{landini:2000} also used Zipfs law when
he looked for meaning in the Voynich manuscript. The consensus of all
researchers is that the emergence of a Zipf relation between phenomena
in a language-like construct does not prove that the item under
consideration is a language, but that a real langua almost certainly
displays Zipf behaviour.

Ferrer and Sol\'e \cite{ferrer01tworegimes} describe a double law
for Zipf, i.e. the fact that the Zipfian curve is best described by
two or even more functions. This suggests the existence of two
'regimes' in english, one general lexicon (5000 for BNC) and a
specialized lexicon.

In a later paper\cite{cancho/sole:2003}, Ferrer and Sol\'e formulate
an attractive model that establishes Zip's law as a necessary traject
in the relation between signifiant and signifi\'e that lies at the
basis of all communication. They use modelling of signals and objects
in a simple binary matrix $a$ of $n$ signals and $m$ objects
(Saussure's {\em signifiant} and {\em signifi\'e}). If a signal
refers to an object, the corresponding cell is one, else it is a
zero.

\begin{figure}
\begin{footnotesize}
\begin{center}
\begin{tabular}{|l| |l|l|l| |l|l|l|}
\hline
                & bank  & stoel  & oever & bank & stoel & oever\\
\hline
furniture       & 1     & 0      & 0     & 0    & 1     & 0    \\
money institute & 1     & 0      & 0     & 1    & 0     & 0    \\
river bank      & 1     & 0      & 0     & 0    & 0     & 1    \\
\hline
&\multicolumn{3}{c}{Least effort for speaker} &\multicolumn{3}{c}{Least effort for listener}\\
\hline
\end{tabular}
\caption{Different efforts for speaker and listener}
\label{banken}
\end{center}
\end{footnotesize}
\end{figure}

In figure \ref{banken} there are two examples of such signal-object
matrices. The english speaker invests no effort when he wants to refer to the
furniture, money institute or river bank; the word 'bank' covers all
three. The listener however has to work very hard to extract the true meaning
from context. The dutch situation is the opposite: the speaker has to select
one from three words, whereas the listener knows immediatly what is referred
to.  

Ferrer and Sol\'e vary the zero's and ones in the matrix using an
evolutionary algorithm. They then use the signal entropy to compute
the minimum cost of the combined effort for both parties in such
matrices:

\[
\Omega(\lambda)=\lambda H_{m}(R|S) + (I-\lambda) H_{n}(S)
\]

where the $\lambda$ parameter weighs the contribution of each party. The
following equation describes the mutual information for different values of
$\lambda$:

\[
I_{n}(S,R)=H_{n}(S)-H_{n}(S|R)
\]
\noindent
and $L$ describes the effective lexicon size in relation to the number of
signals

\[
L=\frac{|\{i|\mu > 0\}|}{n}
\] 
\noindent
where $\mu_{i}=\sum_{j}a_{ij}$ in matrix $a$.

\begin{figure}
\begin{center}
\epsfig{file=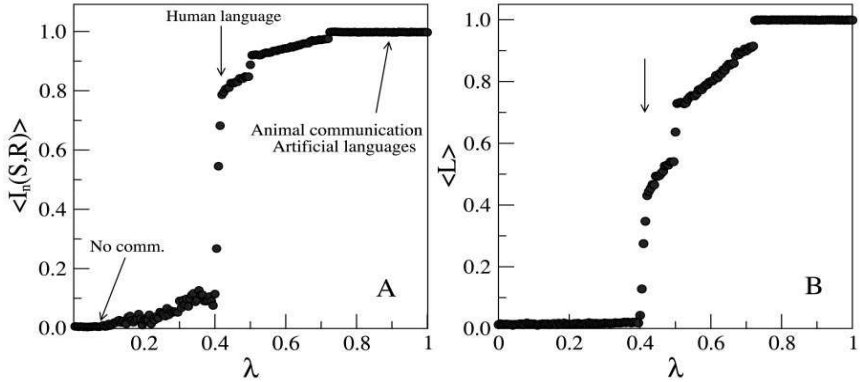,height=2in,width=4.5in}
\caption{\em Relation between  (from \cite{cancho/sole:2003})}
\label{ferrer1}
\end{center}
\end{figure}

If the $I_{n}(S|R)$ and $L$ are plotted against $\lambda$ we get the graphs as
in figure \ref{ferrer1}. It is immediatly clear that there is a catastrophic
transition at $\lambda \approx 0.41$ both for $I_{n}=(S|R)$ and $L$. In a
second experiment, Ferrer and Sol\'e plotted the normalized frequencies of the
signals in the matrices against their rank for different values of
$\lambda$. It was found that Zipf's law emerged in a small window near
$\lambda=0.41$. Together it means that Zipf's law is not just a trivial
outcome of a simple process, as would be suggested by the fact that it is also
valid for random languages, but that it is an intrinsic part of the mathematic
model of communication.

Considering the graphs of figure \ref{ferrer1}, we see that animal
communication is placed in the upper right, where one-to-one signal-object
maps are situated. As we have already seen, this is not the case for the
dolphins and squirrel monkeys of McCowan and possibly other animal systems.

\section{Conclusions}

Corpus Linguistics is the discipline that studies language(groups) from big
samples of that language(group), with a strong emphasis on quantitative
phenomena and methods. Traditionally, of course, such language samples were
restricted to human languages, but with progressive research in biology and
animal communication systems, corpora of non-human language-like phenomena are
a distinct possibility. The sounds emitted by dolphins and collected by
e.g. McCowan \cite{Cowan/Hanser/Doyle:2002,mccowan:1999} clearly constitute a
corpus, and so may other registrations of animal behaviour constitute corpora.

Our survey so far of animal communication centered on the mathematical
qualities of the data, such as Zipfian distribution and entropy, and we sought
to find these qualities in the bee dance data.  Our main problem was and is
therefore whether the bee dance language data can be analysed and stored in
such a way that a Zipfian distribution (if present) can be detected. Partly
this depends on the quantity of the data-types. If the assumptions of Haldane
and Spurway are correct, and if there really are 12 bits of information
contained in the bee dance, this may well be the case. The second and as yet
unsolved problem is the articulation of the data, i.e. the splitting into
meaningfull 'words', and we hope to tackle this problem in the next few months.


\section{Acknowlegdments}

My thanks go to Dr. W.J. Boot from the Laboratory of Entomology,
Wageningen University, Netherlands, and Dr. M. Beekman from Behaviour and
Genetics of Social Insects Lab, School of Biological Sciences, University of
Sydney, Australia. Also to my collegue Drs. M. Reynaert from the institute of
Language and Artificial Intelligence, Tilburg University, Netherlands.

No Microsoft software was used in research or the preparation of the article.

\bibliographystyle{plain}
\bibliography{cuba}

\begin{thebibliography}{10}

\bibitem{havlin-can}
A.~Cohen, R.N. Mantegna, and S.~Havlin.
\newblock Can zipf analyses and entropy distinguish between artificial and
  natural language texts?, 1996.

\bibitem{Dyer:2002}
F.C. Dyer.
\newblock Biology of the dance language.
\newblock {\em Annual Review of Entomology}, 47:917--949, 2002.

\bibitem{haldane:1954}
J.~Haldane and H.~Spurway.
\newblock A statistical analysis of communication in apis mellifera and a
  comparison with communication in other animals.
\newblock {\em Insectes sociaux}, 1:247--283, 1954.

\bibitem{ferrer01tworegimes}
Ramon~Ferrer i~Cancho and Ricard~V. Solé.
\newblock Two regimes in the frequency of words and the origins of complex
  lexicons: Zipf's law revisited.
\newblock {\em Journal of Quantitative Linguistics}, 8(3):165--173, 2001.

\bibitem{cancho/sole:2003}
Ramon~Ferrer i~Cancho and Ricard~V. Solé.
\newblock Least effort and the origins of scaling in human language.
\newblock {\em Proceedings of the national academie of science of the USA},
  2003.

\bibitem{kak:1991}
S.~C. Kak.
\newblock The honey bee dance language controversy.
\newblock {\em The mankind quarterly}, pages 357--365, 1991.

\bibitem{kornai:2002}
A.~Kornai.
\newblock How many words are there?
\newblock {\em Glottometrics}, 4:61--86, 2002.

\bibitem{landini:2000}
G.~Landini.
\newblock Zipf's laws in the voynich manuscript, 2000.

\bibitem{landini:2001}
G.~Landini.
\newblock Evidence of linguistic structure in the voynich manuscript using
  spectral analysis.
\newblock {\em Cryptologia}, XXV:275--295, 2001.

\bibitem{li92random}
Wentian Li.
\newblock Random texts exhibit zipf's law-like word frequency distribution.
\newblock {\em IEEETIT: IEEE Transactions on Information Theory}, 38, 1992.

\bibitem{mccowan:1999}
B.~McCowan, Sean Hanser, and L.~Doyle.
\newblock Quantitative tools for comparing animal communication systems:
  information theory applied to bottlenose dolphin whistle repertoires.
\newblock {\em Animal behaviour}, 57:409--419, 1999.

\bibitem{Cowan/Hanser/Doyle:2002}
Brenda McCowan, Sean Hanser, and Laurance Doyle.
\newblock Using information theory to assess the diversity, complexity and
  development of communicative repertoires.
\newblock {\em Journal of comparative psychology}, 116(2):166--172, 2002.

\bibitem{Reznikova/Ryabko:2001}
Zhanna Reznikova and Boris Ryabko.
\newblock A study of ants' numerical competence.
\newblock {\em Electronic Transactions on Artificial Intelligence}, 5:111--126,
  2001.

\bibitem{Towne:1988}
W.F. Towne and J.L. Gould.
\newblock The spatial precision of the honeybee's dance communication.
\newblock {\em Journal of insect Behaviour}, 1:129--155, 1988.

\bibitem{Ujhelyi:1996}
Maria Ujhelyi.
\newblock Is there any intermediate stage between animal communication and
  language?
\newblock {\em Journal of theoretical Biology}, 180:71--76, 1996.

\bibitem{Frisch:1947}
K.~von Frisch.
\newblock The dance of the bees.
\newblock {\em Bulletin of Animal Behaviour}, 5:1--32, 1947.

\bibitem{wenner/wells:1973}
A.~Wenner and Ph. Wells.
\newblock Do bees have a language?
\newblock {\em Nature}, 241:171--174, 1973.

\bibitem{wenner:2002}
A.M. Wenner.
\newblock The elusive bee dance "language" hypothesis.
\newblock {\em Journal of insect behavior}, 15(6):859--878, November 2002.

\end{thebibliography}


\end{document}